\documentclass[11pt,a4paper]{article}

\usepackage[utf8]{inputenc}
\usepackage[T1]{fontenc}
\usepackage[english]{babel}

\usepackage{geometry}
\usepackage{setspace}
\usepackage{parskip}
\usepackage{titlesec}
\usepackage{enumitem}
\usepackage{needspace}

\usepackage{amsmath,amssymb,amsfonts}
\usepackage{mathtools}
\usepackage{graphicx}
\usepackage{float}
\usepackage{caption}
\usepackage{booktabs}
\usepackage[table]{xcolor}
\usepackage{fancyhdr}
\usepackage{hyperref}
\usepackage{microtype}

\hypersetup{
	colorlinks=true,
	linkcolor=black,
	urlcolor=black,
	citecolor=black,
	pdftitle={IMEX: Interaction-Based Model Explanation}
}

\titleformat{\section}{\Large\bfseries}{\thesection}{1em}{}
\titleformat{\subsection}{\large\bfseries}{\thesubsection}{1em}{}
\titleformat{\subsubsection}{\normalsize\bfseries}{\thesubsubsection}{1em}{}

\titlespacing*{\section}{0pt}{1.2em}{0.6em}
\titlespacing*{\subsection}{0pt}{1.0em}{0.4em}
\titlespacing*{\subsubsection}{0pt}{0.8em}{0.2em}

\let\oldsubsection\subsection
\renewcommand{\subsection}{\Needspace{10\baselineskip}\oldsubsection}
\let\oldsubsubsection\subsubsection
\renewcommand{\subsubsection}{\Needspace{8\baselineskip}\oldsubsubsection}

\setlist[itemize]{leftmargin=1.4cm, itemsep=0.25em, topsep=0.35em}
\setlist[enumerate]{leftmargin=1.4cm, itemsep=0.25em, topsep=0.35em}

\title{
	\textbf{IMEX} \\[0.3em]
	\textbf{Interaction-Based Model Explanation}
}
\author{
	Emiliano Massi\\
	Backwell Tech Corp. Europe GmbH\\
	R\&D Department
}

\date{April 16, 2026}

\begin{document}

	\maketitle

	\begin{abstract}
		In predictive modeling, the ability to explain why a model produces a given target prediction has become increasingly important \cite{doshi2017towards,lipton2018mythos}. Black-box models do not provide a transparent description of the internal mechanisms that generate the prediction, making even accurate predictions difficult to interpret and validate.

		In critical contexts, predictive accuracy alone is not a sufficient validation metric if the reasons underlying model decisions remain unexplained. The IMEX (\textit{Interaction-Based Model Explanation}) approach represents a methodological direction within explainable predictive modeling.

		IMEX is designed to identify which variables contribute most to the target prediction and which interactions among variables are significant in determining the target. The method does not impose limitations on higher-order interaction analysis, allowing the investigation of feature subsets with cardinality greater than two.

		Beyond the identification of feature importance, IMEX enables the exploration of interaction patterns that may be consistent with latent mechanisms influencing the outcome. Through the application of the IMEX algorithm, it is possible to construct an interpretability map of the predictions.

		The IMEX framework is built on two complementary metrics: Static Correlation Power (PCS), which quantifies the contribution of individual features, and Interaction Correlation Power (PCI), which captures non-additive effects among features. In the present work, the PCS component is experimentally validated through a comparison with INVASE \cite{yoon2019invase} on three synthetic datasets with known structures. The results indicate that IMEX can recover relevant feature-level structures in the presence of non-linear, conditional, and multicollinear relationships between input features and prediction targets. The PCI component is introduced as a principled methodological extension of the framework and provides the basis for future interaction-aware benchmarking, particularly with methods such as SHAP interaction values \cite{lundberg2020local}.
	\end{abstract}

	\section{Introduction}

	Machine Learning and Deep Learning models are often characterized by the inability to provide a clear interpretation of their predictions \cite{lipton2018mythos,doshi2017towards}. In critical and high-impact applications, this limitation generates uncertainty, instability, and significantly reduces the possibility of adopting artificial intelligence solutions.

	This challenge has motivated the development of \textit{Explainable Artificial Intelligence} (XAI), whose goal is to design methods capable of making black-box model decisions more transparent by highlighting the mechanisms underlying predictions \cite{ribeiro2016trust,baehrens2010explain,sundararajan2017axiomatic}.

	IMEX is positioned in this context as a \textit{post-hoc} explainability approach. Starting from the predictions generated by a trained model (XGBoost in the proposed experiments \cite{chen2016xgboost}), IMEX constructs an interpretability structure based on scores associated with both individual features and their interactions.

	Feature relevance in the determination of the target is evaluated through two main metrics:

	\begin{itemize}
		\item \textbf{PCS (Static Correlation Power):} measures the individual effect of each feature.
		\item \textbf{PCI (Interaction Correlation Power):} measures the combined effect of interactions among features.
	\end{itemize}

	The present paper focuses primarily on the empirical validation of PCS, which constitutes the experimentally tested component of the framework. PCI is included as a coherent methodological extension that generalizes IMEX toward interaction-aware explainability and higher-order dependency analysis.

	In highly critical contexts, where prediction quality is decisive, understanding the reasons behind predictions increases trust in the system and enables more robust validation. For this reason, a method that is able to recover relevant explanatory drivers under known and controlled data-generation schemes has direct methodological value for XAI.

	\section{Related Work}

	Explainable Artificial Intelligence has produced multiple families of methods aimed at interpreting black-box predictions. These methods differ substantially in their explanatory logic: some identify which input variables are relevant for prediction, others decompose the prediction into additive contributions, and others attempt to quantify interaction effects among variables \cite{molnar2022interpretable,lipton2018mythos}.

	\subsection{Feature Importance and Feature Selection Methods}

	A first relevant family of approaches focuses on identifying the most important variables for a given prediction. In this area, INVASE is particularly relevant because it provides instance-wise variable selection through a neural-network-based framework \cite{yoon2019invase}. From the perspective of the present work, INVASE is the most natural benchmark for PCS because both methods operate at the feature-importance level and both produce one relevance score per feature.

	In this sense, PCS and INVASE are structurally comparable since they both define explanatory outputs in the same feature space $\mathbb{R}^{N}$. This makes their comparison methodologically appropriate in the context of synthetic datasets with known ground truth, where feature-level precision, recall, and F1 score can be computed directly. More generally, this perspective is aligned with model-agnostic work on variable importance and perturbation-based influence measures \cite{fisher2019all,datta2016algorithmic,hooker2019please}.

	\subsection{Additive Explanation Methods}

	A second important family of XAI methods is based on additive attribution. LIME \cite{ribeiro2016trust} and SHAP \cite{lundberg2017unified} are among the most widely used examples. These approaches explain a prediction by decomposing it into the sum of feature-level contributions with respect to a baseline or expected prediction.

	Although SHAP provides a rigorous and influential attribution framework \cite{lundberg2017unified,covert2020global}, its standard feature values are not directly equivalent to PCS. The reason is methodological: PCS measures the observed predictive variation induced by removing a feature from the predictive process, whereas SHAP feature values quantify the additive contribution attributed to that feature within a decomposition of the prediction. Therefore, PCS and SHAP feature values are not directly comparable as numerical objects, even though both concern feature-level relevance.

	For this reason, the present paper considers INVASE as the primary benchmark for PCS. The comparison is more direct because both methods are designed to identify relevant features for prediction rather than to decompose the prediction into additive attribution terms.

	\subsection{Interaction-Based Explanation Methods}

	When the goal is not only to identify relevant variables but also to quantify their joint effects, interaction-based explainability methods become more appropriate. In this context, SHAP interaction values \cite{lundberg2020local} provide an important reference because they extend additive attribution to pairwise interaction effects between features. Related work on ANOVA-style decompositions, Sobol indices, and feature interaction attribution further highlights the relevance of moving beyond single-feature explanations \cite{hooker2007generalized,sobol2001global,owen2014sobol,janizek2021explaining}.

	\subsection{Positioning of IMEX}

	IMEX capable of operating both at the level of individual feature importance, through the PCS (Static Correlation Power) metric, and at the level of feature interaction importance, through the PCI (Interaction Correlation Power) metric.

From the perspective of IMEX, SHAP interaction values are the most natural future benchmark for PCI. At the pairwise level, both PCI and SHAP interaction values operate on the interaction space defined by all unordered feature pairs, that is, a space of dimension $\binom{N}{2}$. This means that they are structurally comparable as interaction descriptors, even though they are derived from different explanatory principles.

More specifically, SHAP interaction values arise from an additive decomposition of the prediction into main effects and pairwise interaction effects, whereas PCI is constructed from non-additive predictive variations obtained through joint feature removal. Therefore, the two approaches are comparable at the level of interaction representation, but they are not numerically equivalent by construction. This is analogous to the distinction between PCS and standard SHAP feature values at the feature level.

	\section{Method}

	Let a dataset with $N+1$ features be given. We select the $ k-th$ target feature and $N$ input features. We define:
	
	\begin{itemize}
		\item $y_k$: the $k-th$ target feature prediction obtained using all features ($N$) (baseline prediction),
		\item $y_{-i,k}$: the $k-th$ target feature prediction  obtained after removing feature $i$,
		\item $y_{-i-j,k}$: the $k-th$ target feature prediction  obtained after removing features $i$ and $j$.
	\end{itemize}

	\subsection{Static Correlation Power (PCS)}

	The variation induced by removing feature $i$ is defined as:

	\begin{equation}
		\Delta y_{i,k} = \left| y_{-i,k} - y_k \right|
	\end{equation}

	The PCS score for feature $i$ is then defined as:

	\begin{equation}
		PCS_i = \frac{\Delta y_{i,k}}{\sum_{j=1}^{N} \Delta y_{j,k}}
	\end{equation}

	Thus, $PCS_i$ quantifies the proportion of total absolute predictive variation attributable to the removal of feature $i$. By construction, PCS provides a normalized feature-level importance measure that can be directly compared across the features of the same prediction task.

	\subsection{Interaction Correlation Power (PCI)}

	For a pair of features $(i,j)$, we define:

	\begin{equation}
		\Delta y_{i,j,k} = \left| y_{-i-j,k} - y_k \right|
	\end{equation}

	\begin{equation}
		I_{ij} = \left| \Delta y_{i,j,k} - \left( \Delta y_{i,k} + \Delta y_{j,k} \right) \right|
	\end{equation}

	Once the interaction factor $I_{ij}$ between features $i$ and $j$ has been computed, we define the \textit{Interaction Correlation Power} associated with the pair $(i,j)$ as a normalized interaction score.

	\begin{equation}
		PCI_{ij} =
		\frac{I_{ij}}
		{
			\left(
			\sum_{\substack{t=1 \\ t \neq i}}^{N} I_{it}
			+
			\sum_{\substack{t=1 \\ t \neq j}}^{N} I_{tj}
			- I_{ij}
			\right)
		}
	\end{equation}

	The denominator represents the total interaction contribution associated with features $i$ and $j$, obtained by:
	\begin{itemize}
		\item summing all interaction terms involving feature $i$ with every other feature,
		\item summing all interaction terms involving feature $j$ with every other feature,
		\item removing the duplicated contribution $I_{ij}$, which would otherwise be counted twice.
	\end{itemize}

	This normalization ensures that $PCI_{ij}$ quantifies the relative importance of the interaction between features $i$ and $j$ with respect to all interactions in which the two features are involved.

	By construction, $PCI_{ij}$ highlights non-additive effects and allows the identification of feature pairs whose joint contribution to the prediction cannot be explained by their individual effects alone.

	The collection of all $PCI_{ij}$ values defines an interaction-level interpretability structure over the feature space. In the current paper, PCI is introduced as a principled extension of the IMEX framework, while its full empirical validation against dedicated interaction-based explainability methods is left for future work.

	\subsection{Generalization to Higher-Order Interactions}

The IMEX framework can be naturally extended to analyze higher-order interactions involving more than two features.

Let $S \subseteq \{1, \dots, N\}$ be a subset of features with cardinality $|S| = m$, where $2 \leq m \leq N-1$.

\begin{equation}
	\Delta y_{S,k} = \left| y_{-S,k} - y_k \right|
\end{equation}

\begin{equation}
	I_S = \left| \Delta y_{S,k} - \sum_{i \in S} \Delta y_{i,k} \right|
\end{equation}

This quantity measures the non-additive contribution of the joint removal of the feature set $S$, capturing interaction effects that cannot be explained by the sum of individual feature contributions.

\vspace{0.3cm}

\begin{equation}
	PCI_S =
	\frac{I_S}
	{
		\sum_{\substack{T \subseteq \{1,\dots,N\} \\ |T| = m \\ T \cap S \neq \emptyset}} I_T
	}
\end{equation}

where the denominator represents the total interaction mass over all subsets $T$ of cardinality $m$ that share at least one feature with $S$.

\vspace{0.3cm}

In practical terms, the normalization can be interpreted as follows:
\begin{itemize}
	\item the numerator isolates the interaction strength of the specific subset $S$,
	\item the denominator aggregates all interaction contributions involving at least one feature in $S$,
	\item this ensures that $PCI_S$ quantifies the relative importance of the joint interaction within its local interaction neighborhood.
\end{itemize}

\vspace{0.3cm}

This formulation extends the IMEX interpretability framework from pairwise interactions to higher-order structures, enabling the detection of complex collective behaviors among features.

In particular:
\begin{itemize}
	\item for $|S| = 2$, the formulation reduces to the pairwise PCI,
	\item for $|S| > 2$, the method captures higher-order dependencies that may reveal patterns consistent with the influence of latent mechanisms not observable at lower interaction orders.
\end{itemize}

\vspace{0.3cm}

\begin{equation}
	\text{Number of subsets of size } m = \binom{N}{m}
\end{equation}

This combinatorial growth suggests that efficient sampling or approximation strategies may be required for large-scale applications.

\vspace{0.3cm}

More generally, the IMEX framework can be extended beyond pairwise interactions to capture dependency structures that emerge only when multiple variables act jointly. In this context, higher-order interaction patterns should be interpreted as collective mechanisms rather than isolated effects.

IMEX does not identify latent variables directly, but higher-order interaction patterns may be consistent with their potential influence.

This aspect is particularly relevant for future developments of the methodology, especially in domains where explanatory mechanisms are known to be collective rather than purely individual. In this sense, PCI and its higher-order extensions, together with the construction of a connection map, define one of the main methodological frontiers of the IMEX framework.

In the current work, this generalization is introduced as a theoretical extension of the IMEX framework, while its systematic empirical validation is left as a direction for future research.

	\section{Experiments}

	IMEX was tested on three synthetic datasets specifically designed to include linear, non-linear, conditional, and multivariate relationships among the features. The experiments were carried out by selecting different prediction targets and comparing IMEX with INVASE \cite{yoon2019invase}.

	The predictive model adopted in the experiments is XGBoost \cite{chen2016xgboost}. This choice is motivated by its ability to capture non-linear relationships, threshold effects, and complex dependencies among features, making it an appropriate model for evaluating post-hoc explainability methods under heterogeneous synthetic scenarios.

	The validity of a post-hoc explainability method depends on the predictive quality of the underlying model. If the predictive model does not adequately capture the data-generating relationships, the explanations derived from it become unstable or weakly informative. For this reason, before evaluating IMEX and INVASE as explainability methods, we report the predictive performance of the XGBoost model on all prediction targets considered in the experiments.

	The explicit functional forms used to generate the datasets are known and controlled but are omitted for brevity, as the evaluation focuses on recovering the structural dependency pattern rather than the exact functional mapping. These rules are coherent with the corresponding connection maps reported for each dataset.

	\begin{table}[H]
		\centering
		\scriptsize
		\resizebox{0.95\textwidth}{!}{
		\begin{tabular}{llrrr}
			\toprule
			\textbf{Dataset} & \textbf{Target} & \textbf{Test MAE} & \textbf{Test RMSE} & \textbf{Test $R^2$} \\
			\midrule
			Dataset 1 & Age & 6.798 & 8.206 & 0.627680 \\
			Dataset 1 & Number of purchases & 0.011 & 0.022 & 0.999751 \\
			Dataset 1 & Spending on promotions & 2.561 & 3.096 & 0.959358 \\
			Dataset 1 & Spending on breakfast & 0.007 & 0.011 & 0.999872 \\
			Dataset 2 & Premium subscription & 0.008 & 0.053 & 0.987324 \\
			Dataset 2 & Annual income & 2231.423 & 2818.151 & 0.949624 \\
			Dataset 2 & Average spending & 6.389 & 7.686 & 0.909174 \\
			Dataset 2 & Annual purchases & 3.989 & 4.963 & 0.564329 \\
			Dataset 3 & Customers per month & 970.031 & 1412.095 & 0.874581 \\
			Dataset 3 & Monthly revenue & 18148.794 & 27364.580 & 0.934417 \\
			Dataset 3 & Operators & 0.562 & 0.712 & 0.960430 \\
			Dataset 3 & Number of pumps & 0.370 & 0.487 & 0.980137 \\
			\bottomrule
		\end{tabular}
		}
		\caption{Predictive performance of the XGBoost model on the targets used in the experiments. Metrics are reported on the test set.}
	\end{table}

	The reported metrics show that the predictive model achieves generally satisfactory to high performance across the considered targets. Most targets are associated with high $R^2$ values, while a limited number of cases are comparatively more challenging. Even in those settings, the model remains sufficiently informative to support a meaningful post-hoc analysis within the controlled synthetic scenarios considered in this paper.

	It is important to distinguish predictive evaluation from explainability evaluation. XGBoost is assessed through predictive regression metrics, whereas IMEX and INVASE are evaluated through their ability to recover the known ground-truth explanatory structure. These are complementary but conceptually distinct validation levels.

	The datasets were designed so that the dependency structure among the variables was known in advance. This makes it possible to compare the explanations generated by the XAI algorithms against an explicit ground truth, following the evaluation logic advocated in the interpretability literature \cite{doshi2017towards}.

	From an experimental standpoint, the present validation concerns the PCS component of IMEX. This choice is methodologically appropriate because PCS produces feature-level importance scores that can be directly compared with INVASE under a controlled ground-truth setting. The PCI component is retained in the paper because it is part of the original framework architecture, but it is not claimed here as a fully benchmarked comparative contribution.

	\section{Results}

	The results section is organized in two parts. First, we describe the comparative evaluation procedure used to assess XAI models. Second, we report the empirical findings obtained on the three synthetic datasets.

	\subsection{Comparative Analysis Procedure for XAI Models}

	In the context of XAI, comparing the outputs produced by different explanation algorithms is particularly relevant for two main reasons:

	\begin{enumerate}[label=\arabic*)]
		\item to determine which XAI model offers better performance under a given scenario described by the input dataset;
		\item to determine, under the same scenario, the degree of complementarity among multiple XAI models.
	\end{enumerate}

	The first analysis is useful for selecting the most appropriate model for a specific context. The second analysis makes it possible to identify a multi-model framework that, if properly combined, can provide a more powerful analysis tool than any single model taken in isolation.

	\subsubsection{Synthetic Datasets with Known Structures}

	Synthetic datasets were generated with predefined relationships among the features. These relationships may be of different types; in our experimentation we considered:

	\begin{itemize}
		\item linear correlations,
		\item multilinear correlations,
		\item inverse proportionality,
		\item conditional relationships.
	\end{itemize}

	For each dataset, a connection map was defined in order to explicitly represent both direct and indirect dependencies among the features.

\subsubsection{Ground Truth Construction}

For each dataset, a prediction target was selected and a corresponding prediction dataset was generated according to the predefined dependency structure.

A \textit{ground truth table} was then built. This table specifies, for each target and for each feature, whether a true explanatory connection is present according to the designed data-generation mechanism. The table is populated with binary values:

\begin{itemize}
	\item $0$: connection absent,
	\item $1$: connection present.
\end{itemize}

Unlike empirical feature relevance, the ground truth used in this work is directly derived from the structural rules governing the synthetic data generation process. This ensures full control over the explanatory mechanisms and allows a rigorous, transparent, and reproducible evaluation of explainability methods.

The ground truth does not represent the predicted target values, but a binary structural map indicating whether each feature is theoretically connected to the target according to the data-generation rules.

In particular, for Dataset 1, the ground truth associated with the prediction target \textit{Promo Spend} is constructed in strict alignment with the connection map and with the dependency rules encoded in the data-generation process. This guarantees that the ground truth represents the true underlying explanatory structure, rather than an inferred or approximated one.

For readability and interpretability purposes, only an illustrative subset of the ground truth table is reported below. The excerpt refers specifically to Dataset 1 (2000 observations). Other datasets have different sizes as reported in their respective sections. The comparative evaluation between IMEX (PCS) and INVASE has been performed on a representative subset of 100 observations.

This subset preserves the structural and statistical properties of the full dataset while allowing a compact and interpretable presentation of the methodology. Since the ground truth is deterministic and defined at the structural level, the use of a subset does not alter the validity of the comparison.

\begin{table}[H]
	\centering
	\scriptsize
	\resizebox{0.9\textwidth}{!}{
		\begin{tabular}{|r|c|c|c|c|c|c|c|}
			\hline
			\textbf{Sample} & \textbf{Age} & \textbf{Household size} & \textbf{N. purchases} & \textbf{Total spend} & \textbf{Time / purchase} & \textbf{Breakfast spend} & \cellcolor{yellow}\textbf{Promo spend} \\
			\hline
			1  & 0 & 1 & 1 & 1 & 0 & 0 & \cellcolor{yellow}0 \\
			2  & 0 & 1 & 1 & 1 & 0 & 0 & \cellcolor{yellow}0 \\
			3  & 0 & 1 & 1 & 1 & 0 & 0 & \cellcolor{yellow}0 \\
			4  & 0 & 1 & 1 & 1 & 0 & 0 & \cellcolor{yellow}0 \\
			5  & 0 & 1 & 1 & 1 & 0 & 0 & \cellcolor{yellow}0 \\
			6  & 0 & 1 & 1 & 1 & 0 & 0 & \cellcolor{yellow}0 \\
			7  & 0 & 1 & 1 & 1 & 0 & 0 & \cellcolor{yellow}0 \\
			8  & 0 & 1 & 1 & 1 & 0 & 0 & \cellcolor{yellow}0 \\
			9  & 0 & 1 & 1 & 1 & 0 & 0 & \cellcolor{yellow}0 \\
			10 & 0 & 1 & 1 & 1 & 0 & 0 & \cellcolor{yellow}0 \\
		\end{tabular}
	}
	\caption{Illustrative excerpt of the ground truth table for Dataset 1 with prediction target \textit{Promo Spend}. The highlighted column corresponds to the target variable. Values equal to 1 indicate the presence of a true explanatory connection according to the predefined data-generation mechanism.}
\end{table}

As shown in the table, the ground truth identifies \textit{Household Size}, \textit{Number of Purchases}, and \textit{Total Spend} as structurally relevant features for the prediction of \textit{Promo Spend}, while \textit{Age}, \textit{Time per Purchase}, and \textit{Breakfast Spend} are not part of the explanatory mechanism for this specific target.

This explicit construction of the ground truth is a key element of the experimental design. It enables a direct and objective comparison between the true explanatory structure and the outputs generated by different XAI methods, allowing a rigorous assessment of their ability to recover meaningful feature relevance.

\subsubsection{Feature-Level Thresholds}

The baseline importance level for the feature scores was computed, and a test threshold was defined as the baseline increased by $20\%$:

\begin{equation}
	\text{Feature threshold} = 1.20 \cdot (\text{baseline})
\end{equation}

This threshold was used to transform continuous importance scores into binary outputs.

The choice of a threshold strictly greater than the baseline is theoretically motivated. If the threshold were set equal to the baseline, all features would be treated as equally relevant under a uniform importance assumption, preventing any meaningful distinction between explanatory and non-explanatory variables.

By setting the threshold above the baseline, only features whose contribution exceeds the uniform-importance condition are retained. This allows the method to focus on structurally significant signals while filtering out weak or non-informative contributions.

It is important to note that the choice of a $20\%$ increase over the baseline does not represent a rigid or universally optimal assumption. Rather, it reflects a principled and balanced trade-off between sensitivity and selectivity in the identification of relevant features. The threshold is not derived from an optimization procedure but from a theoretical consistency argument ensuring deviation from uniform importance.

Lower thresholds would increase sensitivity but may introduce noise, while higher thresholds would enforce stricter selection criteria at the cost of potentially discarding weaker yet meaningful signals. From this perspective, the threshold can be interpreted as a tunable parameter controlling the strength of the comparison.

A systematic exploration of threshold sensitivity and its impact on explainability performance is left for future work.

This binarization step is applied consistently to both PCS and INVASE outputs. In this way, the resulting binary importance tables can be directly compared with the ground truth table, ensuring methodological coherence, interpretability, and fairness in the evaluation process.

	\subsubsection{Importance Table}

	Feature-importance XAI algorithms were then applied to the prediction dataset. For each observation, the corresponding feature-importance scores were obtained. Using the threshold defined above, the continuous output scores were converted into a binary \textit{importance table} according to the following rule:

	\begin{table}[H]
		\centering
		\begin{tabular}{ll}
			\toprule
			\textbf{Condition} & \textbf{Output} \\
			\midrule
			Model feature importance $\geq$ threshold & 1 \\
			Model feature importance $<$ threshold & 0 \\
			\bottomrule
		\end{tabular}
		\caption{Binarization rule for model importance outputs.}
	\end{table}

	In this way, the XAI model output becomes directly comparable with the ground truth table.

	\subsubsection{Explainability Metrics}

	The quality of the explanations is evaluated by comparing:

	\begin{itemize}
		\item the \textit{importance table} generated by the XAI model;
		\item the \textit{ground truth table}.
	\end{itemize}

	This comparison is quantified using standard classification metrics, here interpreted as explainability metrics:

	\begin{itemize}
		\item \textbf{Precision Truth},
		\item \textbf{Recall Truth},
		\item \textbf{F1 Score Truth}.
	\end{itemize}

	These metrics measure how well the XAI model identifies the truly relevant features, both in terms of correctness and completeness. Model comparison was primarily based on the \textit{F1 Score Truth}. In the current study, the benchmarked output of IMEX is the PCS score, which is evaluated as a feature-level explanatory signal against the known ground truth.

	\paragraph{Precision Truth}
	\begin{equation}
		P = \frac{\begin{array}{c}\text{Number of features identified as important by the model}\\ \text{and confirmed by the ground truth}\end{array}}
		{\text{Number of features identified as important by the model}}
	\end{equation}

	\paragraph{Recall Truth}
	\begin{equation}
		R = \frac{\text{Number of features identified by the model and confirmed by the ground truth}}
		{\text{Number of truly important features in the ground truth}}
	\end{equation}

	\paragraph{F1 Score Truth}
	\begin{equation}
		F1 = \frac{2RP}{R+P}
	\end{equation}

	\subsection{Empirical Evaluation on Synthetic Datasets}

	\subsubsection{Dataset 1: Purchase in Store}

	Dataset 1 describes a \textit{purchase in store} use case.

	\begin{itemize}
		\item Number of columns: 7
		\item Number of rows: 2000
	\end{itemize}

	\begin{table}[H]
		\centering
		\begin{tabular}{ll}
			\toprule
			\textbf{Feature} & \textbf{Description} \\
			\midrule
			X1 & Age \\
			X2 & Household size \\
			X3 & Number of purchases \\
			X4 & Total spend \\
			X5 & Time per purchase \\
			X6 & Breakfast spend \\
			X7 & Promotional spend \\
			\bottomrule
		\end{tabular}
		\caption{Dataset 1 features.}
	\end{table}

	\definecolor{mapblue}{RGB}{197,217,241}
	\definecolor{mapyellow}{RGB}{255,255,0}
	\definecolor{mapgreen}{RGB}{146,208,80}
	\definecolor{mapred}{RGB}{255,0,0}

	\begin{table}[H]
		\centering
		\begin{tabular}{|l|l|l|l|l|l|l|}
			\hline
			\textbf{X1} & \textbf{X2} & \textbf{X3} & \textbf{X4} & \textbf{X5} & \textbf{X6} & \textbf{X7} \\
			\hline
			&  & \cellcolor{mapblue} &  &  &  & \cellcolor{mapblue} \\
			\hline
			& \cellcolor{mapblue} & \cellcolor{mapblue} &  &  &  &  \\
			\hline
			&  & \cellcolor{mapyellow} & \cellcolor{mapyellow} &  &  &  \\
			\hline
			& \cellcolor{mapyellow} &  &  &  &  & \cellcolor{mapyellow} \\
			\hline
			& \cellcolor{mapyellow} &  & \cellcolor{mapyellow} &  &  &  \\
			\hline
			\cellcolor{mapgreen} &  &  &  &  & \cellcolor{mapgreen} &  \\
			\hline
			&  &  &  &  &  &  \\
			\hline
			&  & \cellcolor{mapred} &  & \cellcolor{mapred} &  &  \\
			\hline
		\end{tabular}
		\caption{Dataset 1 connection map.}
	\end{table}

	\begin{table}[H]
		\centering
		\begin{tabular}{|l|c|}
			\hline
			Inverse Proportionality & \cellcolor{mapred} \\
			\hline
			Strong Linear Correlation & \cellcolor{mapyellow} \\
			\hline
			Moderate Linear Correlation & \cellcolor{mapblue} \\
			\hline
			``If--Then'' Conditional & \cellcolor{mapgreen} \\
			\hline
		\end{tabular}
		\caption{Dataset 1 mathematical connection types.}
	\end{table}

	\begin{table}[H]
		\centering
		\small
		\begin{tabular}{|l|r|r|r|r|r|r|r|}
			\hline
			& \textbf{X1} & \textbf{X2} & \textbf{X3} & \textbf{X4} & \textbf{X5} & \textbf{X6} & \textbf{X7} \\
			\hline
			X1 & 1 &  &  &  &  & \cellcolor{mapgreen} & \\
			\hline
			X2 & $-0.04877$ & 1 &  &  &  &  & \\
			\hline
			X3 & 0.021622 & \cellcolor{mapblue}$-0.63102$ & 1 &  & \cellcolor{mapred} &  & \\
			\hline
			X4 & $-0.03992$ & \cellcolor{mapyellow}0.825374 & \cellcolor{mapyellow}$-0.87893$ & 1 &  &  & \\
			\hline
			X5 & $-0.03374$ & \cellcolor{mapyellow}0.782247 & \cellcolor{mapyellow}$-0.93329$ & \cellcolor{mapyellow}0.942783 & 1 &  & \\
			\hline
			X6 & \cellcolor{mapyellow}$-0.80423$ & 0.049127 & $-0.00282$ & 0.027931 & 0.01984 & 1 & \\
			\hline
			X7 & $-0.0446$ & \cellcolor{mapyellow}0.982194 & \cellcolor{mapblue}$-0.62121$ & \cellcolor{mapyellow}0.813179 & \cellcolor{mapyellow}0.769066 & 0.048124 & 1 \\
			\hline
		\end{tabular}
		\caption{Dataset 1 correlation matrix.}
	\end{table}

	The tables below summarize the comparison between INVASE and IMEX in terms of \textit{F1 Score Truth}. In this setting, the feature-importance score generated by IMEX is represented by the PCS metric. The results for Dataset 1 are especially informative because they show that PCS remains informative across heterogeneous dependency types, including inverse proportionality, conditional rules, and multicollinear structures.

	\begin{table}[H]
		\centering
		\small
		\begin{tabular}{|l|r|r||l|r|r|}
			\hline
			\multicolumn{3}{|c||}{\cellcolor{mapgreen}\textbf{TARGET: AGE}} & \multicolumn{3}{c|}{\cellcolor{mapgreen}\textbf{TARGET: PURCHASE NUMBER}} \\
			\hline
			\cellcolor{orange}\textbf{FEATURES} & \cellcolor{orange}\textbf{INVASE} & \cellcolor{orange}\textbf{PCS} & \cellcolor{orange}\textbf{FEATURES} & \cellcolor{orange}\textbf{INVASE} & \cellcolor{orange}\textbf{PCS} \\
			\hline
			Number of purchases & 0 & 0 & Age & 0 & 0 \\
			\hline
			Household size & 0 & 0 & Household size & 0 & 0 \\
			\hline
			Total spend & 0 & 0 & Total spend & \cellcolor{mapyellow}0.163636 & \cellcolor{mapyellow}0.126984 \\
			\hline
			Promotional spend & 0 & 0 & Promotional spend & 0 & 0 \\
			\hline
			Breakfast spend & 0 & \cellcolor{mapgreen}0.503704 & Breakfast spend & 0 & 0 \\
			\hline
			Time per purchase & 0 & 0 & Time per purchase & \cellcolor{mapred}{0.586207} & \cellcolor{mapred}{0.545455} \\
			\hline
		\end{tabular}
		\caption{Dataset 1 test results -- Table 1.}
	\end{table}

	\begin{table}[H]
		\centering
		\small
		\begin{tabular}{|l|r|r||l|r|r|}
			\hline
			\multicolumn{3}{|c||}{\cellcolor{mapgreen}\textbf{TARGET: PROMO SPEND}} & \multicolumn{3}{c|}{\cellcolor{mapgreen}\textbf{TARGET: BREAKFAST SPEND}} \\
			\hline
			\cellcolor{orange}\textbf{FEATURES} & \cellcolor{orange}\textbf{INVASE} & \cellcolor{orange}\textbf{PCS} & \cellcolor{orange}\textbf{FEATURES} & \cellcolor{orange}\textbf{INVASE} & \cellcolor{orange}\textbf{PCS} \\
			\hline
			Age & 0 & 0 & Age & \cellcolor{mapgreen}1 & \cellcolor{mapgreen}0.979798 \\
			\hline
			Number of purchases & 0 & \cellcolor{mapyellow}0.471698 & Number of purchases & 0 & 0 \\
			\hline
			Household size & 0 & \cellcolor{mapyellow}0.458015 & Household size & 0 & 0 \\
			\hline
			Total spend & \cellcolor{mapyellow}1 & \cellcolor{mapyellow}0.458015 & Total spend & 0 & 0 \\
			\hline
			Breakfast spend & 0 & 0 & Promotional spend & 0 & 0 \\
			\hline
			Time per purchase & 0 & 0 & Time per purchase & 0 & 0 \\
			\hline
		\end{tabular}
		\caption{Dataset 1 test results -- Table 2.}
	\end{table}

	\paragraph{Comments on Dataset 1}
	\begin{itemize}
		\item In the case of inverse proportionality between target and features, the two methods detect the connection and show substantial agreement.
		\item In the case of conditional relationships, the methods are again in agreement, although INVASE does not detect the bidirectional nature of the relationship, whereas IMEX signals this condition, even if at a lower level.
		\item In the presence of multicollinearity between target and features, PCS appears to recover a broader portion of the explanatory structure. This is visible, for example, for the target \textit{Promo Spend}, where INVASE highlights one main correlation while PCS captures a pattern that is more consistent with the predefined structure.
	\end{itemize}

	\clearpage
	\subsubsection{Dataset 2: Purchase Online}

	Dataset 2 describes a \textit{purchase online} use case.

	\begin{itemize}
		\item Number of columns: 9
		\item Number of rows: 5000
	\end{itemize}

	\begin{table}[H]
		\centering
		\begin{tabular}{ll}
			\toprule
			\textbf{Feature} & \textbf{Description} \\
			\midrule
			X1 & Age \\
			X2 & Annual income \\
			X3 & Annual purchases \\
			X4 & Average spend \\
			X5 & Premium subscription \\
			X6 & Number of complaints \\
			X7 & Days since last purchase \\
			X8 & Devices used \\
			X9 & Time on platform \\
			\bottomrule
		\end{tabular}
		\caption{Dataset 2 features.}
	\end{table}

	\begin{table}[H]
		\centering
		\begin{tabular}{|l|l|}
			\hline
			\textbf{RANGE -- R} & \textbf{LEVEL} \\
			\hline
			0.9 -- 1 & \cellcolor{mapred}{EXTREME} \\
			\hline
			0.7 -- 0.89 & \cellcolor{mapyellow}HIGH \\
			\hline
			0.50 -- 0.69 & \cellcolor{mapblue}MODERATE \\
			\hline
			0.30 -- 0.49 & \cellcolor{mapyellow!30}LOW \\
			\hline
			0.00 -- 0.29 & NO SIGN \\
			\hline
		\end{tabular}
		\caption{Dataset 2 absolute correlation level.}
	\end{table}

	\begin{table}[H]
		\centering
		\small
		\resizebox{\textwidth}{!}{%
		\begin{tabular}{|l|r|r|r|r|r|r|r|r|r|}
			\hline
			& \textbf{X1} & \textbf{X2} & \textbf{X3} & \textbf{X4} & \textbf{X5} & \textbf{X6} & \textbf{X7} & \textbf{X8} & \textbf{X9} \\
			\hline
			X1 & 1 &  &  &  &  &  &  &  & \\
			\hline
			X2 & \cellcolor{mapred}{0.915998} & 1 &  &  &  &  &  &  & \\
			\hline
			X3 & \cellcolor{mapyellow}$-0.75682$ & \cellcolor{mapblue}$-0.694673$ & 1 &  &  &  &  &  & \\
			\hline
			X4 & \cellcolor{mapyellow}0.873483 & \cellcolor{mapred}{0.9558943} & \cellcolor{mapblue}$-0.66085$ & 1 &  &  &  &  & \\
			\hline
			X5 & \cellcolor{mapyellow}0.723123 & \cellcolor{mapyellow}0.7841456 & \cellcolor{mapblue}$-0.54721$ & \cellcolor{mapyellow}0.752224 & 1 &  &  &  & \\
			\hline
			X6 & $-0.20257$ & $-0.224171$ & 0.150027 & $-0.20836$ & $-0.28549$ & 1 &  &  & \\
			\hline
			X7 & 0.247064 & 0.2222666 & \cellcolor{mapyellow!30}$-0.32706$ & 0.212705 & 0.176731 & $-0.03284$ & 1 &  & \\
			\hline
			X8 & 0.075767 & 0.0978838 & $-0.05019$ & 0.090907 & 0.030485 & $-0.02461$ & 0.020031 & 1 & \\
			\hline
			X9 & $-0.08791$ & $-0.079393$ & 0.139198 & $-0.07609$ & $-0.07333$ & 0.040573 & $-0.09581$ & $-0.01638$ & 1 \\
			\hline
		\end{tabular}%
		}
		\caption{Dataset 2 correlation matrix.}
	\end{table}

	The results for Dataset 2 further support the validity of PCS. In particular, the comparison suggests that the PCS metric is able not only to recover the strongest drivers, but also to detect weaker yet still meaningful explanatory signals that remain partially unobserved by INVASE.

	\begin{table}[H]
		\centering
		\small
		\begin{tabular}{|l|r|r||l|r|r|}
			\hline
			\multicolumn{3}{|c||}{\cellcolor{mapgreen}\textbf{TARGET: PREMIUM SUBSCRIPTION}} & \multicolumn{3}{c|}{\cellcolor{mapgreen}\textbf{TARGET: ANNUAL PURCHASES}} \\
			\hline
			\cellcolor{orange}\textbf{FEATURES} & \cellcolor{orange}\textbf{Invase} & \cellcolor{orange}\textbf{PCS} & \cellcolor{orange}\textbf{FEATURES} & \cellcolor{orange}\textbf{Invase} & \cellcolor{orange}\textbf{PCS} \\
			\hline
			Annual income & 0 & 0.071429 & Premium subscription & 0 & \cellcolor{mapblue}0.52 \\
			\hline
			Devices used & 0 & 0 & Devices used & 0 & 0 \\
			\hline
			Age & 0 & \cellcolor{mapyellow}0.257143 & Age & 0 & \cellcolor{mapyellow}0.59542 \\
			\hline
			Days last purchase & 0 & 0 & Days last purchase & 0 & 0 \\
			\hline
			Number of complaints & 0 & 0.181818 & Number of complaints & 0 & 0 \\
			\hline
			Annual purchases & \cellcolor{mapblue}0.726115 & \cellcolor{mapblue}0.391304 & Annual income & \cellcolor{mapblue}0.675497 & \cellcolor{mapblue}0.409639 \\
			\hline
			Average spend & 0 & \cellcolor{mapyellow}0.625 & Average spend & 0 & 0 \\
			\hline
			Time on platform & 0 & 0 & Time on platform & 0 & 0 \\
			\hline
		\end{tabular}
		\caption{Dataset 2 test results -- Table 1.}
	\end{table}

	\begin{table}[H]
		\centering
		\small
		\begin{tabular}{|l|r|r||l|r|r|}
			\hline
			\multicolumn{3}{|c||}{\cellcolor{mapgreen}\textbf{TARGET: ANNUAL INCOME}} & \multicolumn{3}{c|}{\cellcolor{mapgreen}\textbf{TARGET: AVERAGE SPEND}} \\
			\hline
			\cellcolor{orange}\textbf{FEATURES} & \cellcolor{orange}\textbf{Invase} & \cellcolor{orange}\textbf{PCS} & \cellcolor{orange}\textbf{FEATURES} & \cellcolor{orange}\textbf{Invase} & \cellcolor{orange}\textbf{PCS} \\
			\hline
			Premium subscription & 0 & \cellcolor{mapyellow}0.508475 & Premium subscription & 0 & \cellcolor{mapyellow}0.530973 \\
			\hline
			Annual purchases & 0 & 0 & Annual purchases & 0 & 0 \\
			\hline
			Devices used & 0 & 0 & Devices used & 0 & 0 \\
			\hline
			Age & \cellcolor{mapred}{0.329114} & \cellcolor{mapred}{0.477273} & Age & 0 & 0 \\
			\hline
			Days last purchase & 0 & 0 & Days last purchase & 0 & 0 \\
			\hline
			Number of complaints & 0 & 0 & Number of complaints & 0 & 0 \\
			\hline
			Average spend & \cellcolor{mapred}{0.507463} & \cellcolor{mapred}{0.466667} & Annual income & \cellcolor{mapred}{0.675497} & \cellcolor{mapred}{0.614035} \\
			\hline
			Time on platform & 0 & 0 & Time on platform & 0 & 0 \\
			\hline
		\end{tabular}
		\caption{Dataset 2 test results -- Table 2.}
	\end{table}

	\paragraph{Comments on Dataset 2}
	\begin{enumerate}[label=\alph*)]
		\item In the presence of strong correlations, INVASE and IMEX are in agreement in identifying the main target drivers.
		\item On this dataset, PCS appears more sensitive in detecting smaller but still meaningful correlations, thereby providing a richer explanatory picture in these settings.
	\end{enumerate}

	\clearpage
	\subsubsection{Dataset 3: Gas Station}

	Dataset 3 describes a \textit{gas station} use case.

	\begin{itemize}
		\item Number of columns: 8
		\item Number of rows: 6000
	\end{itemize}

	\begin{table}[H]
		\centering
		\begin{tabular}{ll}
			\toprule
			\textbf{Feature} & \textbf{Description} \\
			\midrule
			X1 & Number of pumps \\
			X2 & Area \\
			X3 & Operators \\
			X4 & Gasoline price \\
			X5 & Monthly revenue \\
			X6 & Customers per month \\
			X7 & Distance to shopping center (km) \\
			X8 & Distance to city center (km) \\
			\bottomrule
		\end{tabular}
		\caption{Dataset 3 features.}
	\end{table}

	\begin{table}[H]
		\centering
		\begin{tabular}{|l|c|}
			\hline
			``If--Then'' Conditional & \cellcolor{mapgreen} \\
			\hline
		\end{tabular}
		\caption{Dataset 3 mathematical connection types.}
	\end{table}

	\begin{table}[H]
		\centering
		\begin{tabular}{|l|l|l|l|l|l|l|l|}
			\hline
			\textbf{X1} & \textbf{X2} & \textbf{X3} & \textbf{X4} & \textbf{X5} & \textbf{X6} & \textbf{X7} & \textbf{X8} \\
			\hline
			&  &  &  & \cellcolor{mapgreen} & \cellcolor{mapgreen} &  &  \\
			\hline
			&  &  &  & \cellcolor{mapgreen} &  & \cellcolor{mapgreen} &  \\
			\hline
			\cellcolor{mapgreen} &  &  &  &  &  & \cellcolor{mapgreen} &  \\
			\hline
			\cellcolor{mapgreen} &  & \cellcolor{mapgreen} &  &  &  &  &  \\
			\hline
			\cellcolor{mapgreen} & \cellcolor{mapgreen} &  &  &  &  &  &  \\
			\hline
		\end{tabular}
		\caption{Dataset 3 connection map.}
	\end{table}

	The results for Dataset 3 are especially informative because they suggest that PCS remains capable of detecting explanatory structure even when the dependency mechanism is conditional and indirect rather than purely correlational.

	\begin{table}[H]
		\centering
		\small
		\begin{tabular}{|l|r|r||l|r|r|}
			\hline
			\multicolumn{3}{|c||}{\cellcolor{mapgreen}\textbf{TARGET: CUSTOMER PER MONTH}} & \multicolumn{3}{c|}{\cellcolor{mapgreen}\textbf{TARGET: MONTHLY REVENUE}} \\
			\hline
			\cellcolor{orange}\textbf{FEATURES} & \cellcolor{orange}\textbf{Invase} & \cellcolor{orange}\textbf{PCS} & \cellcolor{orange}\textbf{FEATURES} & \cellcolor{orange}\textbf{Invase} & \cellcolor{orange}\textbf{PCS} \\
			\hline
			Dist.\ to shopping center & 0 & 0 & Customers per month & \cellcolor{mapgreen}0.83908 & \cellcolor{mapgreen}0.756757 \\
			\hline
			Dist.\ to city center & 0 & 0 & Dist.\ to shopping center & 0 & \cellcolor{mapgreen}0.586207 \\
			\hline
			Monthly revenue & \cellcolor{mapgreen}0.768293 & \cellcolor{mapgreen}0.6 & Dist.\ to city center & 0 & 0 \\
			\hline
			Number of pumps & 0 & \cellcolor{mapgreen}0.5 & Number of pumps & 0 & 0 \\
			\hline
			Operators & 0 & 0 & Operators & 0 & 0 \\
			\hline
			Gasoline price & 0 & 0 & Gasoline price & 0 & 0 \\
			\hline
			Area & 0 & 0 & Area & 0 & 0 \\
			\hline
		\end{tabular}
		\caption{Dataset 3 test results -- Table 1.}
	\end{table}

	\begin{table}[H]
		\centering
		\small
		\begin{tabular}{|l|r|r||l|r|r|}
			\hline
			\multicolumn{3}{|c||}{\cellcolor{mapgreen}\textbf{TARGET: NUMBER OF PUMPS}} & \multicolumn{3}{c|}{\cellcolor{mapgreen}\textbf{TARGET: OPERATORS}} \\
			\hline
			\cellcolor{orange}\textbf{FEATURES} & \cellcolor{orange}\textbf{Invase} & \cellcolor{orange}\textbf{PCS} & \cellcolor{orange}\textbf{FEATURES} & \cellcolor{orange}\textbf{Invase} & \cellcolor{orange}\textbf{PCS} \\
			\hline
			Customers per month & 0 & 0 & Customers per month & 0 & 0 \\
			\hline
			Dist.\ to shopping center & 0 & 0 & Dist.\ to shopping center & 0 & \cellcolor{mapgreen}0.166667 \\
			\hline
			Dist.\ to city center & 0 & 0 & Dist.\ to city center & 0 & 0 \\
			\hline
			Monthly revenue & 0 & 0 & Monthly revenue & \cellcolor{mapgreen}0.913978 & \cellcolor{mapgreen}0.479339 \\
			\hline
			Operators & 0 & \cellcolor{mapgreen}0.571429 & Number of pumps & 0 & \cellcolor{mapgreen}0.794872 \\
			\hline
			Gasoline price & 0 & 0 & Gasoline price & 0 & 0 \\
			\hline
			Area & \cellcolor{mapgreen}1 & \cellcolor{mapgreen}0.721519 & Area & 0 & \cellcolor{mapgreen}0.258621 \\
			\hline
		\end{tabular}
		\caption{Dataset 3 test results -- Table 2.}
	\end{table}

	\paragraph{Comments on Dataset 3}
	On this dataset, PCS appears more sensitive than INVASE in identifying conditional connections, particularly in scenarios characterized by indirect and structured dependencies.

	\subsection{Summary of Results}

	Overall, the experimental comparison suggests that IMEX:

	\begin{itemize}
		\item provides a PCS-based feature-importance signal that is aligned with the known ground truth on these datasets,
		\item shows more consistent behavior than INVASE in some conditions involving non-linear, conditional, and multicollinear structures,
		\item recovers relevant drivers that may remain partially undetected by competing feature-level explainability methods,
		
	\end{itemize}

	From the standpoint of the present paper, these results constitute the main empirical evidence in support of IMEX. More specifically, they validate PCS as a feature-importance metric under controlled synthetic conditions with known explanatory structure.

	\section{Discussion}

	PCS and PCI belong to different interpretability spaces:

	\begin{equation}
		PCS \in \mathbb{R}^{N}, \qquad PCI \in \mathbb{R}^{\binom{N}{2}}
	\end{equation}

	This means that the two quantities are not directly comparable in absolute numerical terms. To address this issue, we introduce a baseline level for each interpretability space:

	\begin{equation}
		Baseline_{PCS} = \frac{1}{N}, \qquad Baseline_{PCI} = \frac{1}{\binom{N}{2}}
	\end{equation}

	Interpretability should therefore be regarded as a \textit{relative} concept, measured with respect to the corresponding baseline. A feature or an interaction is relevant not because of its raw numerical value alone, but because of its position relative to the uniform-importance condition defined by the baseline.

	This perspective supports a structured interpretation of the outputs:
	\begin{itemize}
		\item high PCS values indicate strong individual drivers,
		\item high PCI values indicate relevant interaction mechanisms,
		\item combined PCS--PCI analysis may reveal structured explanatory patterns not observable through feature-level inspection alone.
	\end{itemize}

	The central experimental result of this work concerns PCS. The comparative analysis against INVASE \cite{yoon2019invase} on three synthetic datasets with known dependency structures shows that PCS is able to recover relevant explanatory features with consistent behavior across heterogeneous settings. This is especially important because the datasets were intentionally designed to include not only strong linear dependencies, but also inverse relationships, conditional rules, and multicollinear effects.

	From a methodological perspective, PCS is directly comparable with INVASE because both methods define one importance value per feature and therefore operate in the same feature space $\mathbb{R}^{N}$. This makes the adopted comparison scientifically coherent. By contrast, PCS is not fully comparable to standard SHAP feature values \cite{lundberg2017unified}, even though both concern feature-level relevance. SHAP feature values arise from an additive decomposition of the prediction, whereas PCS is based on the observed predictive variation induced by removing a feature from the model input. The two methods therefore address related but not identical explanatory questions.

	At the interaction level, the situation is analogous rather than contradictory. PCI and SHAP interaction values \cite{lundberg2020local} are not numerically equivalent, but they share the same interaction space and are therefore structurally comparable when pairwise interactions are considered. This makes SHAP interaction values a natural reference point for future PCI benchmarking. At the same time, the two approaches remain different in their explanatory construction: SHAP interaction values arise from an attribution-based additive decomposition, while PCI is derived from non-additive predictive variations induced by joint feature removal.

	PCI should therefore be interpreted in this paper as a methodological extension of the IMEX framework rather than as a fully benchmarked empirical component. Its formal construction is coherent with the PCS rationale and provides a natural mechanism to isolate non-additive effects between features. This makes PCI relevant for the study of interaction-driven explanatory mechanisms and for the analysis of patterns that may be consistent with the influence of latent mechanisms, even though latent variables themselves are not directly identified by the method.

	At the same time, a rigorous comparative validation of PCI requires dedicated benchmarks against interaction-oriented methods, such as SHAP interaction values \cite{lundberg2020local}, as well as datasets with explicitly controlled interaction ground truth. For this reason, PCI is included here as an original and structurally important component of IMEX, but its full empirical assessment is left for future research.

	More broadly, the current evidence suggests the following interpretation: IMEX is experimentally validated at the feature-importance level through PCS, while it is methodologically extended toward interaction-aware explainability through PCI. The present results also suggest that IMEX and INVASE should not necessarily be viewed only in competitive terms: depending on the scenario, they may be in agreement, complementary, or differently sensitive to the same explanatory structure.

	\section{Conclusion}

	IMEX represents an explainability framework aimed at analyzing both individual feature contributions and complex interaction mechanisms. The experimental results on synthetic datasets indicate that IMEX can provide informative explanations alongside alternative methods such as INVASE, especially when the relationship between input features and targets is non-linear or conditionally structured.

	In the present work, this empirical claim is specifically supported by the PCS component, which has been benchmarked under controlled ground-truth conditions and has shown strong performance across diverse dependency scenarios. PCI complements this contribution by extending the framework toward interaction-aware explainability and higher-order dependency analysis, although its dedicated empirical validation remains part of future work.

	Future developments will focus not only on the comparative validation of PCI against interaction-based methods such as SHAP interaction values, but also on the construction of a connection map representing the structure of dependencies between features in terms of strength and interaction patterns.

\end{document}